\def\year{2020}\relax
\documentclass[letterpaper]{article} 
\usepackage{aaai20}  
\usepackage{times}  
\usepackage{helvet} 
\usepackage{courier}  
\usepackage[hyphens]{url}  
\usepackage{graphicx} 
\usepackage{graphicx}  
\title{Copy-Move Forgery Classification via Unsupervised Domain Adaptation}

\author{Akash Kumar \\ MANAS Lab, IIT Mandi \\ akash\_bt2k15@dtu.ac.in \And  Arnav Bhavsar \\
MANAS Lab, IIT Mandi \\
arnav@iitmandi.ac.in
}
 
\begin{document}

\maketitle

\begin{abstract}
In the current era, image manipulation is becoming increasingly easier, yielding more natural looking images, owing to the modern tools in image processing and computer vision techniques. The task of the segregation of forged images has become very challenging. To tackle such problems, publicly available datasets are insufficient. In this paper, we propose to create a synthetic forged dataset using deep semantic image inpainting algorithm. Furthermore, we use an unsupervised domain adaptation network to detect copy-move forgery in images.  Our approach can be helpful in those cases, where the classification of data is unavailable. 
\end{abstract}

\section{Introduction}
With advancement of new image editing technologies 
the number of forgery cases is increasing manifold. 
As a result, in recent years, several deep learning algorithms such as Image Region Forgery Detection \cite{Zhang2016ImageRF}, Augment and Adapt \cite{augment} and BusterNet \cite{busternet} have been proposed to counterattack the problem of image forgery. There are diverse ways of forging images, of which Copy-Move and Splicing forgery are the most common ones. In this paper, we mainly focus on Copy-Move forgery (CMF).  

CMF is a type of passive image forgery technique in which a section of an image is copied and pasted within the same image. Many post-image processing operations such as rescaling, affine transformations, resizing, and blurring can be applied to the copied region. 
As the source and target image remains the same, the photometric characteristics of the image remain largely invariable. Thus, the detection of this type of forgery becomes even more difficult. 

\textbf{Contributions} In this paper, the primary task considered is that of classification of forged and authentic images, for which we employ Unsupervised Domain Adaptation (DA). Moreover, as the publicly available datasets are small, we generate a new dataset of forged images using deep semantic inpainting algorithm from COCO \cite{coco} dataset. To the best of our knowledge, the use of unsupervised learning has not been exploited for the classification of CMF.

\begin{figure}[t]
\centering
\includegraphics[width=0.9\columnwidth]{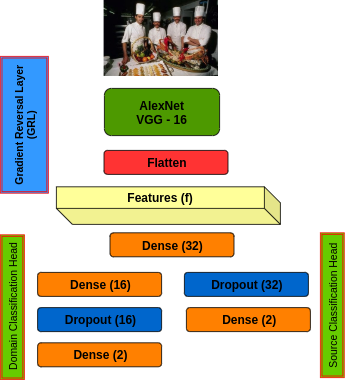} 
\caption{The network comprise of two heads: Source Classification and Domain Classification.}
\label{fig1}
\end{figure}

\begin{figure*}[t]
\centering
\includegraphics[width=0.7\textwidth]{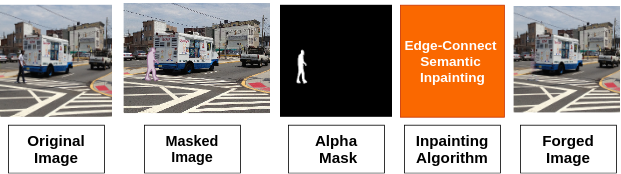} 
\caption{Dataset Generation}
\label{fig2}
\end{figure*}

\section{Approach}
\textbf{Dataset} We synthesized a dataset of CMF images using COCO. We have used all the sub-categories of COCO dataset equally to create a dataset of approximately 15,000 images. The mask of particular sub-categories was cropped out. After that, Edge-connect \cite{edgeconnect} algorithm is used. It is a type of Deep Semantic inpainting. Edge-connect uses a two-stage approach to complete an image. Firstly, the edge generator fills out the missing edges, and then the image completion network completes the image based on the edges deduced.  Using the above approach, we created the training dataset. We evaluated our architecture on CASIA V2 \cite{casia} dataset. It contains 4975 images (Authentic - 1701, Forged - 3274). The method of dataset generation is illustrated in Figure \ref{fig2}.


\textbf{Domain Adaptation} We used Domain Adversarial neural network (DANN) \cite{dann}, a sub-category of Unsupervised DA algorithm.
It creates a feature space of images present in the source and target domain. The distribution of feature representations is such that it is discriminative among various classes and invariant across domains. In our case, COCO dataset is the source domain, and CASIA V2 is the target domain. DANN optimize three parameters during backpropagation: Source classifier (SC) ($\theta$\textsubscript{\textit{s}}), Feature mapping ($\theta$\textsubscript{\textit{f}}) and Domain classifier (DC) ($\theta$\textsubscript{\textit{d}}). The overall loss function is defined in the equation below:

\begin{equation}
    L = L \textsubscript{source} (\theta \textsubscript{\textit{f}}, \theta \textsubscript{\textit{s}}) + L \textsubscript{domain} (\theta \textsubscript{\textit{f}}, \theta \textsubscript{\textit{d}})
\end{equation}{}

DANN has two separate heads. In the SC head, feature mapping and label classifier are optimized as such to reduce the classification loss in case of the source domain. While in DC, feature mapping maximizes the domain loss so that the distribution of both domains becomes similar. It simultaneously minimizes the classification loss for the image, whether it comes from a source or target domain. In this way, the network increases the confusion between source and target domain, so that the model focuses more on the features that help to distinguish images amongst different labels.

\begin{table}[b]
\caption{Accuracy of different architectures on CASIA V2.}\smallskip
\centering
\resizebox{.95\columnwidth}{!}{
\smallskip\begin{tabular}{|l|l|l|}
\hline
\textbf{Baseline Network} & \textbf{Images used} & \textbf{F1-Score}\\ \hline
AlexNet + DANN & 10k  & \textbf{78.8}\\ \hline
VGG-16 + DANN & 10k  & 67.5\\ \hline
BusterNet & 100k & 77.4\\ \hline
\end{tabular}
}
\label{table1}
\end{table}

\textbf{Architecture} The base model of our architecture extracts features from images. Extensive experiments were done using AlexNet \cite{alexnet} and VGG-16 \cite{Simonyan15}  for feature extraction. After feature extraction, the DC network predicts the domain of the input image, and, the SC network predicts the label for source samples. In DANN, there is a particular layer called GRL, which is present in the DC network. During forward pass, it acts as an identity transform. At the time of backpropagation, it multiplies the gradient by a negative constant (-$\lambda$). Figure \ref{fig1} depicts the architecture of our proposed approach. 
At the time of training, we know whether the source domain image is authentic or forged, while we did not use labels of the target domain. We used binary labels to indicate whether the input image comes from the source or target distribution. At the test time, the prediction was made on the whole target domain that is on all 4795 images of CASIA V2 dataset. 

\section{Results and Conclusion}
In this paper, we sugested a new approach for data augmentation to counter the problem of small publicly available datasets for image forgery. We also outlined a novel unsupervised learning approach to detect CMF in images. We presented evaluation over different feature extraction models. Our approach outperforms the accuracy to the previous method, which incorporates supervised deep learning. In the future, we aim to generate more dataset that contains spliced as well as CMF forgeries to make our model more robust. 

\fontsize{9.0pt}{10.0pt} \selectfont

\bibliographystyle{aaai} \bibliography{ref.bib}

\end{document}